\UseRawInputEncoding
\documentclass[letterpaper]{article} 
\usepackage{aaai24}  
\usepackage{times}  
\usepackage{helvet}  
\usepackage{courier}  
\usepackage[hyphens]{url}  
\usepackage{graphicx} 
\def\UrlFont{\rm}  
\usepackage{natbib}  
\usepackage{caption} 
\usepackage{algorithm}
\usepackage{algpseudocode}

\usepackage{amsfonts}
\usepackage{amsmath}
\usepackage{multicol}

\usepackage{newfloat}
\usepackage{listings}
\DeclareCaptionStyle{ruled}{labelfont=normalfont,labelsep=colon,strut=off} 
\floatstyle{ruled}
\newfloat{listing}{tb}{lst}{}
\floatname{listing}{Listing}
\nocopyright

\title{Diversity Measures:\\
Domain-Independent Proxies for Failure in Language Model Queries}
\author{Noel Ngu, Nathaniel Lee, Paulo Shakarian}
\affiliations{Arizona State University\\ Tempe, AZ\\ \{nngu2,nlee51,pshak02\}@asu.edu}

\def\ba{\mathbf{a}}
\def\br{\mathbf{r}}
\def\aset{\mathbb{A}}
\def\indfcn{\mathbf{1}}
\def\calp{\mathcal{P}}
\def\calq{\mathcal{Q}}
\def\prompt{\calp}
\def\ques{\calq}
\def\univ{\mathbb{U}}
\def\reals{\mathbb{R}}
\def\embed{\mathit{embed}}
\def\cent{\mathit{cent}}
\begin{document}

\maketitle

\begin{abstract}
Error prediction in large language models often relies on domain-specific information.  In this paper, we present measures for quantification of error in the response of a large language model based on the diversity of responses to a given prompt - hence independent of the underlying application.  We describe how three such measures - based on entropy, Gini impurity, and centroid distance - can be employed.  We perform a suite of experiments on multiple datasets and temperature settings to demonstrate that these measures strongly correlate with the probability of failure.  Additionally, we present empirical results demonstrating how these measures can be applied to few-shot prompting, chain-of-thought reasoning, and error detection. 
 \end{abstract}

\section{Introduction}
Prompt engineering techniques for large language models (LLMs) have yielded significant results in reducing errors.  Advances such as self-consistency and chain-of-thought prompting~\cite{wei_chain--thought_2022,wang_self-consistency_2022,yao_tree_2023}, in particular, have shown the ability to reduce errors for a given problem.  In this paper, we take a different, but complementary approach to these methods.  Specifically, we study \textit{diversity measures} that can function as a domain-independent proxy for failure probability for question-answer tasks that rely on the prompting of the LLM.  The intuition is that the diversity of the response suggests uncertainty.  In our approach, we avoid requirements of direct access to the LLM~\cite{kuhn2023semantic} as well as semantic understanding (which is often domain-specific)~\cite{lin2023generating,kuhn2023semantic,shakarian_independent_2023}.  Our contributions are as follows.

\begin{itemize}
    \item{We formalize three diversity measures with respect to LLM prompting (Section~\ref{sec:approach} and show, across three datasets and five temperature settings that all three measures correlate closely with the probability of failure with $R^2$ values exceeding $0.8$ in most cases and often higher (Sections~\ref{sec:set-based_eval}-\ref{sec:vec-based_eval}).  We note that these measures do not require access to the LLM or domain knowledge.}
    \item We introduce the concept of \textit{diversity-based prompt selection} (Section~\ref{sec:prompt-sel}).  Experimentally (Section~\ref{sec:promptEng}), we show in two datasets, significant improvement ($14\%,45\%$) over baselines (and near the baseline in a third).  In all cases, this method significantly improved over worst-case prompt outcomes.  
    \item We also extend the results on prompt selection for use with chain-of-thought (CoT) prompting~\cite{wei_chain--thought_2022}.  We show experimentally (Section~\ref{sec:cot}) that CoT and diversity-driven prompting converge to the same failure probability. When combined, failure probability is reduced by $30\%$ as compared to when either technique is used in isolation.
    \item We describe how such measures can be used in machine learning models for predicting failure (Section~\ref{sec:error-pred}) and show experimentally (Section~\ref{sec:error-pred-exp}) that they can be viable for error prediction.  We show how diversity measures can be used in a supervised model to predict errors significantly outperforming the baseline, in the best case obtaining an AUPRC of $0.703$.
\end{itemize}

The rest of the paper is organized as follows.  In Section~\ref{sec:rw} we review related work.  In Section~\ref{sec:approach} we discuss our approach, providing specifics on how we compute entropy, Gini impurity, and centroid-based diversity measures with respect to the output of self-consistency-style prompting~\cite{wang_self-consistency_2022}.  This is followed by a discussion on how the approach is applied to diversity-based prompt selection and error prediction in Section~\ref{sec:app}.  We then present experimental evaluations in Section~\ref{sec:exp} where we study how the diversity measures correlated with failure, diversity-based prompt selection, synergies with CoT prompting, and how such measures can be used with machine learning approaches to predict errors. We provide an implementation available at \url{https://github.com/lab-v2/diversity_measures}
\vspace{5pt}


\section{Related Work}
\label{sec:rw}
There have been numerous recent studies evaluating the performance of LLMs in various settings~\cite{liu_evaluating_2023,shakarian_independent_2023,frieder_mathematical_2023,tu_chatlog_2023,ray_chatgpt_2023,kocon_chatgpt_2023,yuan_how_2023,pardos_learning_2023,openai_gpt-4_2023} these studies compare LLM performance across tasks, models, and datasets.  While these lead to important insights, they generally do not look into the use of specific metrics of the output to improve performance in the general case.  One exception is \cite{shakarian_independent_2023} which studies indicators of ChatGPT failure with respect to math word problems by counting the mathematical operators in the template equations present in the metadata of the DRAW-1K dataset.  Drawing on this idea, a simple ML model was created based on counts of various mathematical operations.  The approach in this paper is not tied specifically to mathematical operations.

Examining the diversity of LLM outputs has also been a topic of previous studies. The selection of tokens using various methods such as temperature, top-$k$, and nucleus sampling~\cite{ficler_controlling_2017,fan_hierarchical_2018,holtzman_curious_2019} to create the output has been studied. These results have generally been integrated into the system design of contemporary LLMs.  Essentially, these findings allow for the selection of tokens that appear in the generated text of an LLM.  As the methods are probabilistic, multiple runs lead to slightly different results.  The aforementioned work provides hyper-parameters to adjust the variance of the LLM response to a given prompt.  Recent work on prompt engineering~\cite{wei_chain--thought_2022,wang_self-consistency_2022,yao_tree_2023} has leveraged this stochastic property to improve results by taking a sample of LLM responses to a given prompt.  However, that work focuses on leveraging a set of LLM outputs to improve accuracy for a given problem, rather than on using such outputs to assess the correctness of the result.

This work is complementary to other active areas of research such as meta-learning~\cite{hospedales_meta-learning_2022,zhou_domain_2022} and introspection~\cite{daftry_introspective_2016,ramanagopal_failing_2018}.  However, these methods often rely on features to predict failure or make a determination of what model to use.  The diversity measures introduced in this paper may enable such techniques to be leveraged in the context of LLMs.

Concurrent with this work there have been several other studies either recently published or in preprint on language model uncertainty.  For example, \cite{zhou2023navigating} found that the presence of words relating to uncertainty in the result of an LLM lead is indicative of errors.  Studies such as \cite{tian2023just} study how to leverage human calibration to create confidence sources for language model results.  In \cite{agrawal2023language} the authors look to evaluate LLM results for hallucinated sources by comparison with external sources.  The work of \cite{kuhn2023semantic} also uses entropy measurements, but relies on access to the token probabilities of the LLM (which we do not assume) and utilizes an additional step of semantic comparison among pairs of language model results - which requires a second supervised model.  The recent preprint of \cite{lin2023generating} builds on that work by creating a second trained model to approximate token probabilities.  In this work, we avoid the training of an additional second model, pairwise comparisons, or access to the original LLM.

\section{Approach}
\label{sec:approach}
\noindent\textbf{Technical Preliminaries.}  Our intuition is to build on the concept of \textit{self-consistency}~\cite{wang_self-consistency_2022}.  For a given prompt-question pair $\prompt, \ques$, the decoder returns a reasoning path, answer pair $\br,\ba$ where $\ba$ is an element of some (possibly infinite sized) answer set $\aset$.  In this paradigm, the decoder is queried $m$ times producing $m$ such $\br,\ba$ pairs.  The final answer is computed based on the answer $a \in \aset$ that maximized the following conditional probability.
\begin{eqnarray}
\label{eqn:selfConExact}
    P(a | \prompt, \ques) = \sum_{i=1}^{m} \indfcn(\ba_i =a)P(\br_i,\ba_i | \prompt, \ques)
\end{eqnarray}
In practice, Equation~\ref{eqn:selfConExact} can be accurately approximated with a majority vote over the $m$ answers (shown in Equation~\ref{eqn:majVote}).  This has the nice property of avoiding the requirement of knowing $P(\br_i,\ba_i | \prompt, \ques)$ - which is not available for most commercial models and is a core assumption in this paper.  Hence, we shall use Equation~\ref{eqn:majVote} as the definition of this probability.
\begin{eqnarray}
\label{eqn:majVote}
    P(a | \prompt, \ques) \approx \frac{1}{m}\sum_{i=1}^{m} \indfcn(\ba_i =a)
\end{eqnarray}

\noindent\textbf{Diversity Measures.}  Our intuition is that given answers $\ba_1,\ldots,\ba_m$ the \textit{diversity} among the answers provides insights into the correctness of the decoder's response.  We consider two paradigms in which to view a given $\ba_i$.  First, for ``set-based'' output we shall assume a universe of elements $\univ$ and that each $\ba_i \subseteq \univ$.  The intuition here is that this view would be suitable for applications such as math and logic problems where the $\ba_i$ is a set of natural numbers or symbols.  In the second paradigm, ``vector-based'' output, we use an embedding function $\embed$ that maps each $\ba_i$ into a $D$ dimensional space ($\embed : \aset \mapsto \reals ^ D$).  This is suitable where a set representation of the output is too coarse-grain or where the reasoning path ($\br_i$) is considered part of the answer itself.\vspace{10pt}

\noindent\textbf{Set-Based Output.} With set-based output, each $\ba_i$ is a set of elements from some universe $\univ$.  For a given element $e \in \univ$, we will define the probability of $e$ appearing in the result as:
\begin{eqnarray}
P(e) = \frac{1}{m}\sum_{i=1}^{m} \indfcn(e \in \ba_i)
\end{eqnarray}
With this quantity, we can directly apply Entropy and Gini impurity measures to create diversity measurements as follows.
\begin{eqnarray}
    H = - \sum_{e \in \univ}P(e)\log(P(e))\\
    G = 1-\sum_{e \in \univ}P(e)^2
\end{eqnarray}


\noindent\textbf{Vector-Based Output.} Here we look to evaluate each $\ba_i$ by examining its vector embedding of dimension $D$.  We will use a distance function $d : \reals^D \times \reals^D \mapsto \reals^+$ to compare two vectors of size $D$.  Given the vectors created by the $\embed$ function $\embed(\ba_1),\ldots,\embed(\ba_m)$ we can calculate a $D$ dimension centroid $\cent$.  Note that we can vary the calculation of $\embed, d,$ and $\cent$ and explore such variations empirically.  Using this information, we propose the following diversity measure.
\begin{eqnarray}
    \frac{1}{m}\sum_{i}d(\embed(a_i),\cent)
\end{eqnarray}

\section{Applications}
\label{sec:app}
In this section, we describe two applications of diversity measures introduced in the previous section: diversity-based prompt selection (Section~\ref{sec:prompt-sel}) and error prediction (Section~\ref{sec:error-pred}).

\subsection{Diversity-Based Prompt Selection}
\label{sec:prompt-sel}
A key problem in developing LLM-powered applications is the design of the prompt.  A diversity measure, such as entropy, can be used to evaluate prompts during training but can also be used in a strategy that employs multiple prompts in what we term \textit{diversity-based prompt selection}.  For example, suppose we have $N$ different prompts, each queried $m$ times.  Using majority voting, we get $N$ responses.  We can then measure each response by calculating the diversity (e.g., entropy) over the $m$ queries associated with each response and take the response with the lowest diversity.

\subsection{Error Prediction}
\label{sec:error-pred}
Following the intuition of work on introspection~\cite{daftry_introspective_2016,ramanagopal_failing_2018}, we look to train a machine learning model in a supervised fashion to predict failures for a given use case.  The input to the model consists of entropy, Gini impurity, and centroid-based features.



\section{Evaluation}
\label{sec:exp}
Using three datasets described in Section~\ref{sec:Setup}, we examine the entropy, Gini impurity, and distance from centroid measures in each of the three datasets and report on their correlation with failure probability in Sections~\ref{sec:set-based_eval} for set-based measures and Section~\ref{sec:vec-based_eval} for vector-based measures.  We then describe experiments on our application results focused on diversity-based prompt selection (Section~\ref{sec:promptEng}) and extend those results for CoT prompting (Section~\ref{sec:cot}).  We then look at the application to failure prediction by describing our supervised model in Section~\ref{sec:error-pred-exp}.

\subsection{Experimental Setup}
\label{sec:Setup}
In all of our experiments, we used OpenAI's GPT 3.5 model for the LLM and interfaced with the LLM through the OpenAI API using Python 3.x.  For experiments on vector-based output, the $\embed$ function was implemented using sentence-BERT~\cite{reimers-gurevych-2019-sentence}.

We evaluated three datasets listed in Table~\ref{tab:ds_tab}.  These datasets were chosen with the intention of examining LLM performance across a diverse set of reasoning tasks (commonsense, arithmetic, and symbolic reasoning.) CSQA consists of $9,741$ samples, DRAW-1K consists of $1,000$ samples, and LL consists of $3,000$ samples.  Splits on training and testing as well as temperature settings are described for individual experiments.

\begin{table}[!ht]
    \centering
    \begin{small}
    \begin{tabular}{|l|l|l|}
    \hline
        {Dataset} & {Domain} & {Response} \\ 
        &&{Type}  \\\hline\hline
         {CSQA} &Knowledge- & Multiple \\
\cite{amrita} &based Q\&A & choice\\
                  \hline
         {DRAW-1K} & Math word  & Numerical \\
          \cite{upadhyay_annotating_2017}& problems & answers\\
         \hline
         {LL} & Last-letter & Textual  \\         \cite{diao2023active}&
          concatenation& concatenation\\
         \hline         
    \end{tabular}
    \end{small}
    \caption{Datasets used in evaluation.}
    \label{tab:ds_tab}
\end{table}

\subsection{Measure Evaluation: Set-Based Output}
\label{sec:set-based_eval}
\begin{figure}
\includegraphics[width=0.5\textwidth]{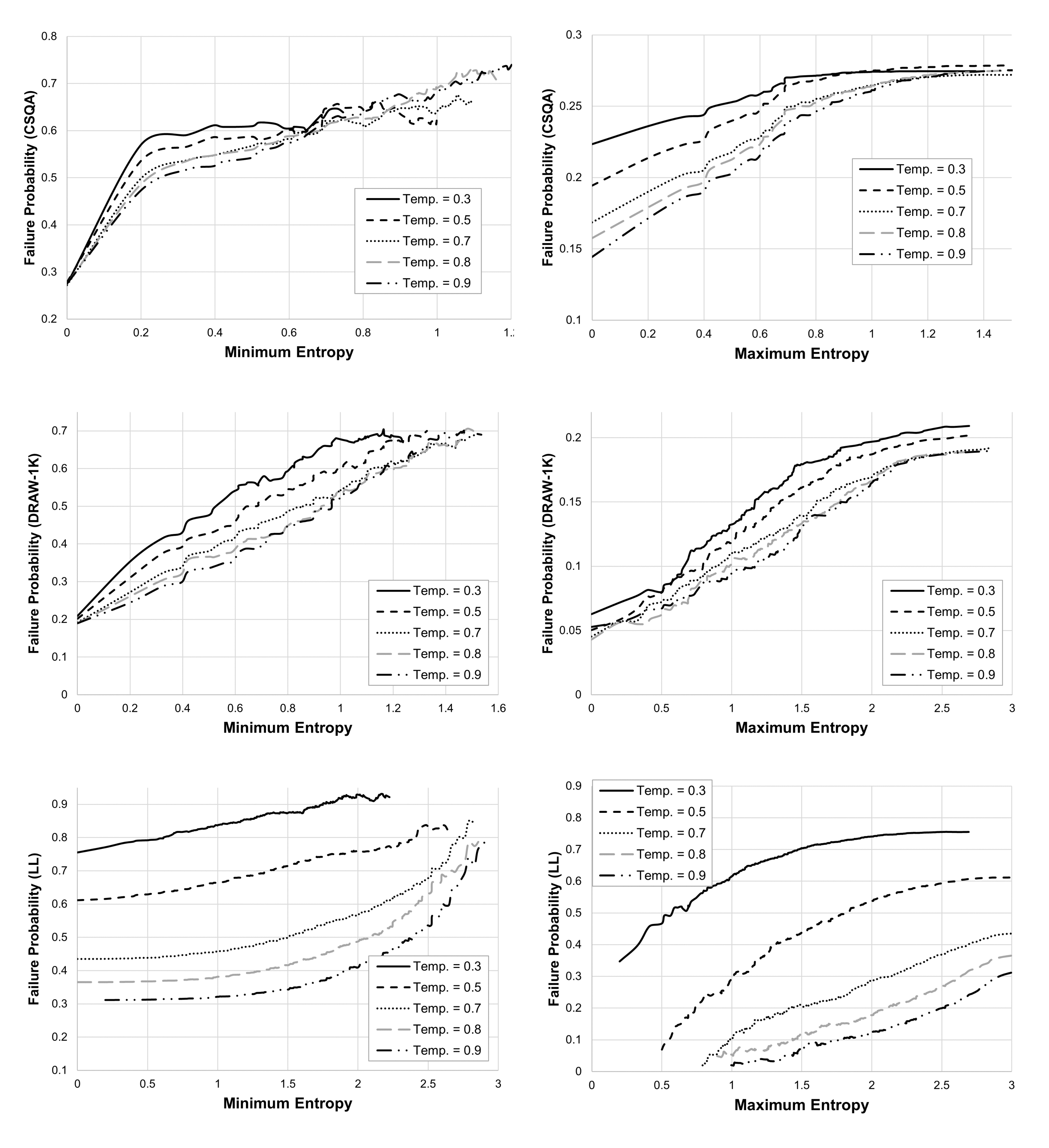}
\caption{Entropy vs. failure probability (top row: CSQA, middle row: DRAW-1K, bottom row: LL)}
\label{fig:entropy}
\end{figure}

\begin{figure}
\includegraphics[width=0.5\textwidth]{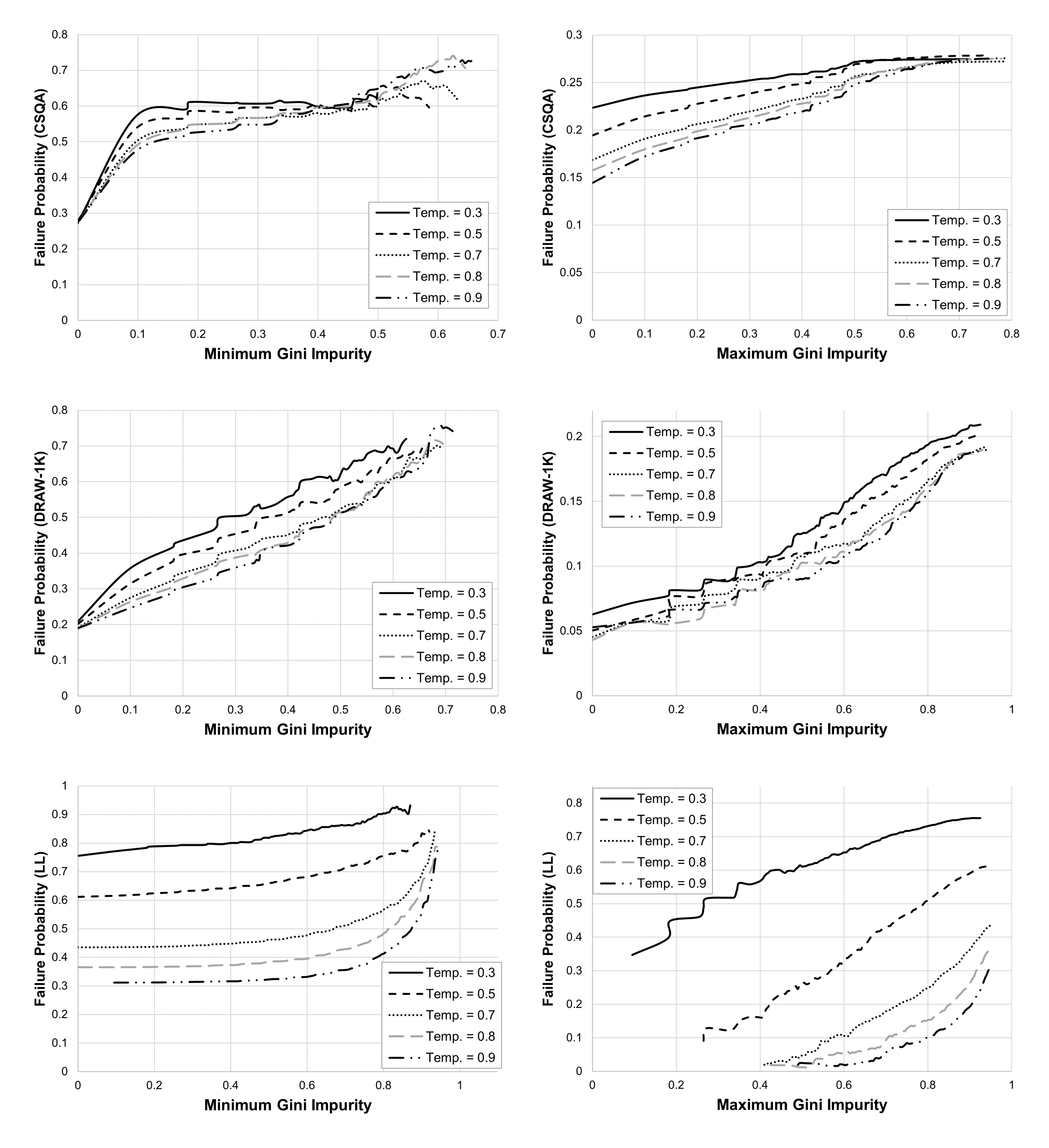}
\caption{Gini impurity vs. failure probability (top row: CSQA, middle row: DRAW-1K, bottom row: LL)}
\label{fig:gini}
\end{figure}

We evaluated entropy and Gini impurity measures on the entirety of each of the three datasets with temperature settings of $\{0.3, 0.5, 0.7, 0.8, 0.9\}$.  We examine cumulative distributions (both for minimum and maximum values) for all values of the measure with a minimum bucket size of 100 samples.  Entropy results are shown in Figure~\ref{fig:entropy} while Gini impurity results are shown in Figure~\ref{fig:gini}.

\noindent\textbf{Entropy Results.}  Entropy measures appeared to be strongly related to the probability of failure for cumulative plots in both directions.  In most cases, the relationship appeared linear with $R^2$ values (for cumulative minimum values with temperature of $0.7$) at $0.808$ for CSQA, $0.973$ for DRAW-1K, and $0.876$ for LL.  Relationships had a tighter linear fit for cumulative plots based on maximum values with $R^2$ values (temperature of $0.7$) of $0.799$ for CSQA, $0.978$ for DRAW-1K, and $0.979$ for LL.

\noindent\textbf{Gini Impurity Results.}  Gini impurity was also closely related to entropy in this case and produced similar trends. We included a plot of the cumulative relationships in Figure~\ref{fig:gini}.

\subsection{Measure Evaluation: Vector-Based Output}
\label{sec:vec-based_eval}
\begin{figure}
\includegraphics[width=0.5\textwidth]{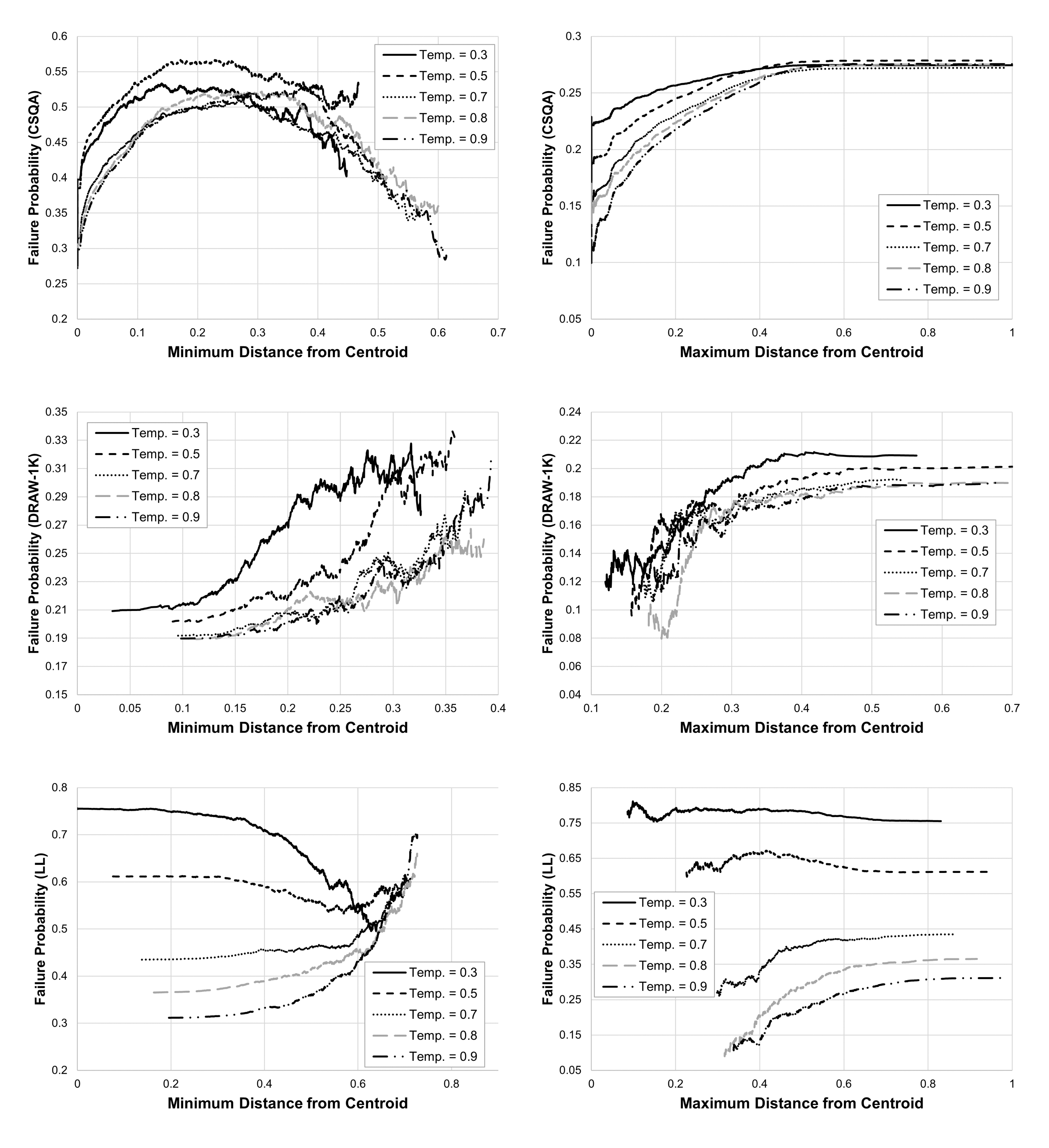}
\caption{Average distance to centroid vs. failure probability (top row: CSQA, middle row: DRAW-1K, bottom row: LL)}
\label{fig:centroid}
\end{figure}

For vector-based output, we evaluated the average distance to the centroid across the same datasets and temperature settings for set-based output (see Section~\ref{sec:set-based_eval} for details).  Cumulative plots are shown in Figure~\ref{fig:centroid}.  In cases where the failure probability is less than $0.5$, we observed a clear monotonic relationship between the average distance to the centroid and failure probability for cumulative plots in both directions.  For cumulative plots based on the maximum average distance from the centroid, the relationship becomes less interesting in CSQA and LL for the temperature settings of $0.3$ and $0.5$ - exhibiting little slope.  For cumulative plots based on the minimum value, we see the trend reversing in the case of LL for these temperature settings - and a similar reversal in the same plots of CSQA across all temperature settings once the maximum failure probability is crested.  This contradicts our original intuition -  that diversity will, in general, lead to uncertainty in the result (and increase the chance of an incorrect response being returned).  However, in these cases of high failure probability and extreme temperature, such diversity may actually lead to a correct answer being produced by chance.  It is notable that this phenomenon is less pronounced in DRAW-1K, likely due to that problem not being multiple choice (making guessing more prone to error).

\subsection{Prompt Selection Experiments}
\label{sec:promptEng}
\begin{figure}
\includegraphics[width=0.5\textwidth]{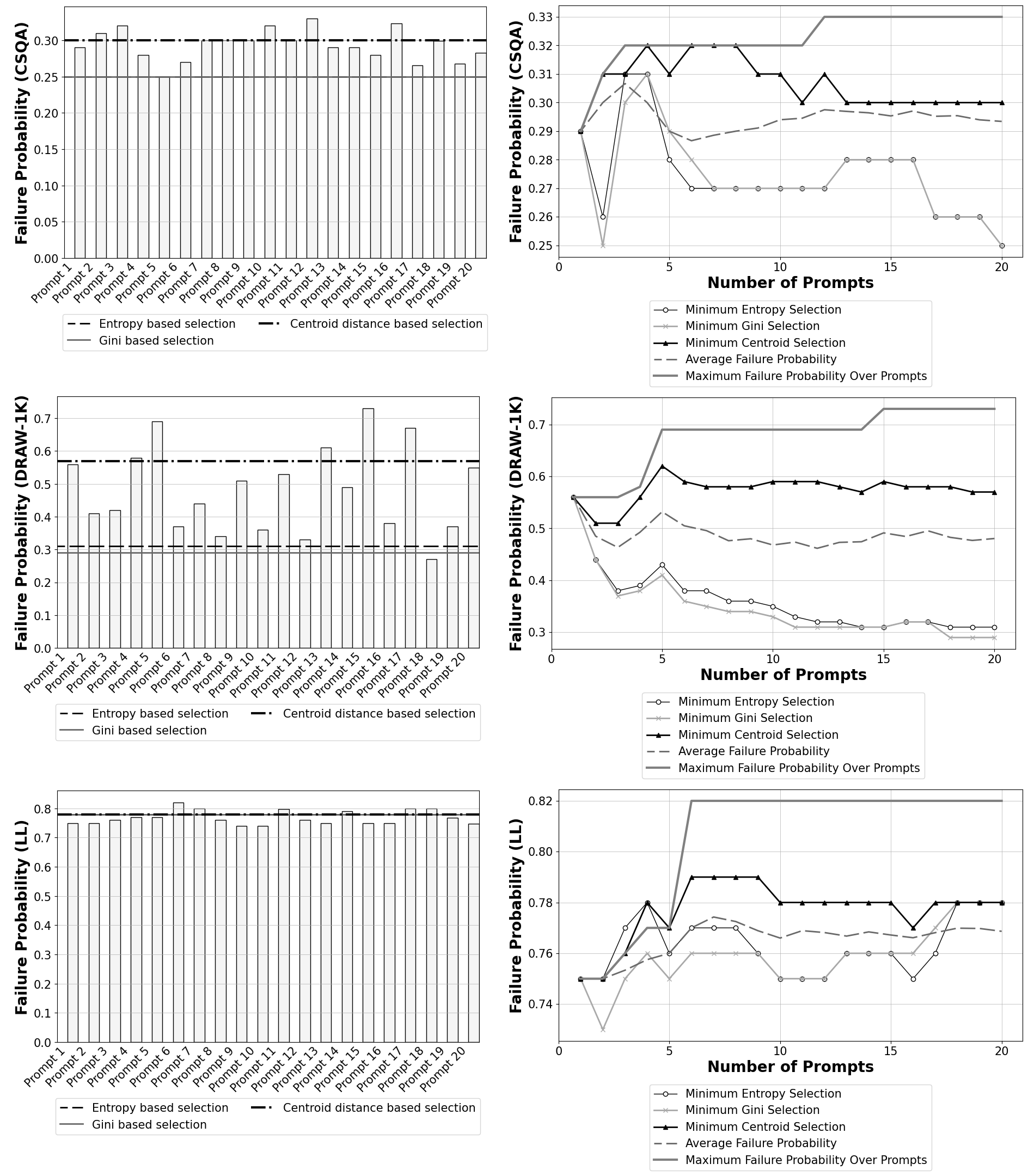}
\caption{Prompt engineering experiment.  Top row: CSQA, middle row: DRAW-1K, bottom row: LL}
\label{fig:fewshot}
\end{figure}

As described earlier, we studied diversity measure's applications to prompt engineering, studying the effects of selecting the prompt providing the lowest entropy response.  Here we generated $20$ prompts from each dataset by randomly selecting $30$ samples with ground-truth responses to serve as few-shot samples for each prompt.  We then randomly selected 100 samples from each dataset, ensuring that these samples were drawn from a population not previously used for any prompt, to create a test set.  Results are shown in Figure~\ref{fig:fewshot}.  The panels on the left show the failure probabilities for each of the $20$ prompts used and compare them with the selection of a prompt based on minimum entropy, Gini impurity, and average distance to the centroid (note for each query, all prompts are used and the one providing the lowest diversity measure is used to return the result).

\noindent\textbf{Entropy and Gini Impurity Prompt Selection Results.}  Comparisons with individual prompts are shown in the left-hand column of Figure~\ref{fig:fewshot}.  Entropy and Gini impurity measures performed comparably.  Selection by both measures outperformed $19$ of the $20$ prompts and obtained the same probability of failure on the one prompt it did not outperform.  On average, it provided a $14\%$ improvement (i.e., an average difference in failure probability of $0.04$) over the CSQA prompts.  Performance was more pronounced with DRAW-1K, which also outperformed $19$ of the $20$ prompts, but on average provided $45\%$ improvement (i.e., an average difference in probability of $0.17$) for entropy, with a slightly better performance obtained with Gini impurity.  Performance with LL was less impressive.  However, unlike the other datasets, the failure probability for all prompts in this experiment exceeded $0.5$.  Here, minimum entropy selection outperformed $6$ of the $20$ prompts, but on average did $1.6\%$ worse (i.e., average increase in probability by $0.01$).  However, in the cases where it did outperform the prompts, it provided a $2.6\%$ improvement - so the approach at least shows it can mitigate against poorly-performing prompts.

We note that the use of $20$ prompts per problem (and each used with $20$ queries) can become an expensive operation, especially considering API and runtime costs.  Therefore, we examined in the right-hand panels of Figure~\ref{fig:fewshot} the failure probability as a function of the number of prompts used.  Note that an average in this case functions as an oracle, as operationally we would not know which prompt to select.  We note that our method (when using entropy or Gini impurity) showed the ability to reduce failure probability below the average without requiring all $20$ prompts, stably reaching a lower failure probability with 5 prompts for CSQA, and 2 prompts for DRAW-1K.  We did not see consistent improvement over average with LL (the only case where failure exceeded $0.5$) but did see the failure probability of minimum entropy selection improve over the worst-case stay within $1.5\%$ of the average with only $5$ prompts.
\vspace{5pt}

\noindent\textbf{Centroid Distance Prompt Selection Results.}  Centroid-based prompt selection consistently outperformed the worst-case scenario, but under-performs (relative to entropy and Gini-based prompt selection) with respect to the average case - in this regard, the result was poorest for DRAW-1K ($18\%$ worse than the average prompt failure probability).  This is likely due to vector representations of numeric answers near each other in the vector space differing significantly in semantics (e.g., a missing negative sign may result in such a phenomenon).

\subsection{CoT Prompting}
\label{sec:cot}
Chain of thought (CoT) prompting~\cite{wei_chain--thought_2022} utilizes explanations in the few-shot examples.  One would suspect that the use of CoT prompting would limit the variety of responses to a prompt, hence limiting the impact of a diversity technique employed to select the correct prompt.  As the DRAW-1K dataset included the associated math equations with ground truth, we used this information to replicate the experiment of Section~\ref{sec:promptEng} with CoT-style prompts.

\begin{figure}
\includegraphics[width=0.5\textwidth]{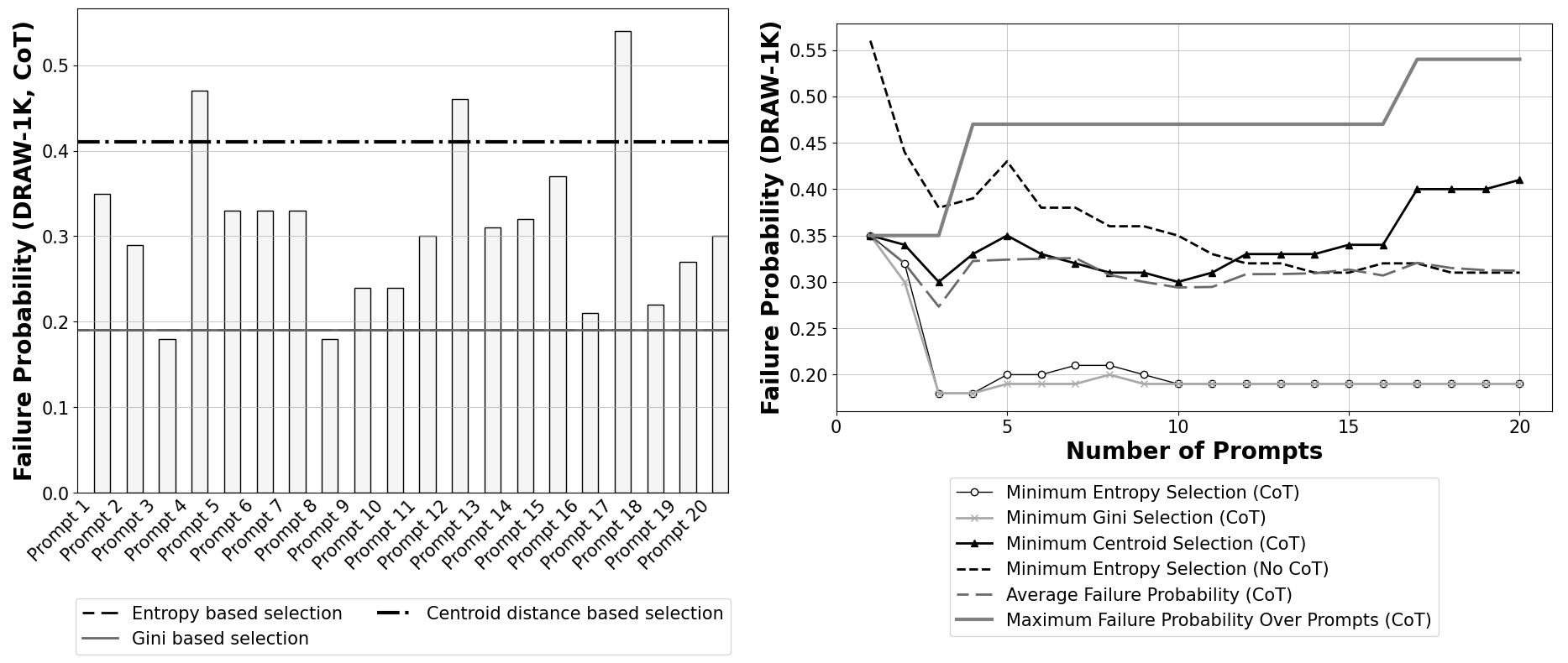}
\caption{Prompt engineering experiment using Chain of Thought (CoT) on DRAW-1K.}
\label{fig:cot}
\end{figure}

\noindent\textbf{Entropy and Gini Impurity Results (with CoT).}  Again, entropy and Gini impurity performed comparably.  First, we observe (in Figure~\ref{fig:cot}, left) that both CoT (without entropy/Gini) and minimum entropy/Gini selection (without CoT) converge to a failure probability of $0.31$ which suggest that both are reducing diversity.

However, we believe that few-shot prompting and CoT reduce diversity in different ways, as there was a synergistic effect from combining them.  Using only $3$ prompts, minimum entropy/Gini selection with CoT consistently provides over a $30\%$ reduction in failure probability over either technique alone.  We also observe the two techniques together provide improved failure probability in $18$ of the $20$ CoT prompts (Figure~\ref{fig:cot}, right).
\vspace{5pt}

\noindent\textbf{Centroid Distance Results (with CoT).}  Again, centroid distance under-performs when compared to entropy and Gini impurity, though combined with CoT provides better results than without.  We also note that DRAW-1K is the dataset for which centroid distance performs the poorest with standard prompt selection and underperformance is likely due to similar issues as described in Section~\ref{sec:promptEng}.

\subsection{Error Prediction Experiments}
\label{sec:error-pred-exp}
\begin{figure}
\centering
\includegraphics[width=0.45\textwidth]{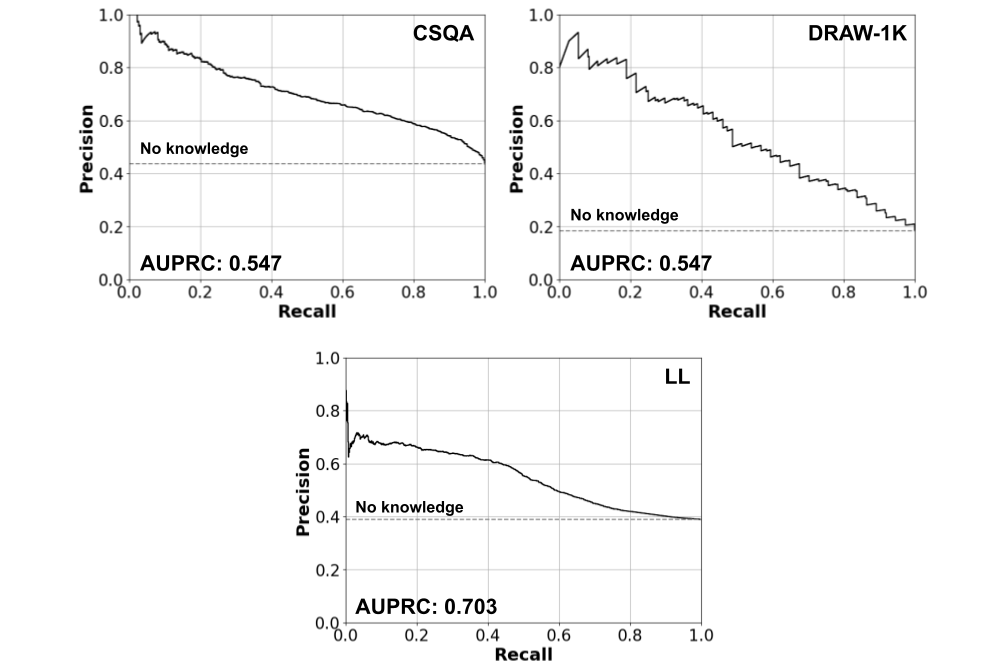}
\caption{Precision-Recall Curves for the MLP-10 failure prediction model CSQA (left), DRAW-1K (center), and LL (right) for temperature=0.7, dashed lines show knowledgeless baseline}
\label{fig:pr-curves}
\end{figure}

We created machine learning models based on the features described in Section~\ref{sec:error-pred} and trained models (specific to each dataset) with a training dataset balanced between cases where the prompt failed and where it was successful. To balance the training data, the majority class was randomly oversampled.  We then evaluated it on a separate test dataset that was not artificially balanced. Precision-recall curves for a 10-layer multi-layer perceptron (implemented in TensorFlow) are shown in Figure~\ref{fig:pr-curves}.  We note that in all cases, the models provide significant improvement in precision over a knowledgless selection.  Further, for low recall setting (under $0.2$) we obtain over $0.7$ precision for DRAW-1K and over $0.8$ precision in LL.
\vspace{5pt}

\begin{figure}
\centering
\includegraphics[width=0.5\textwidth]{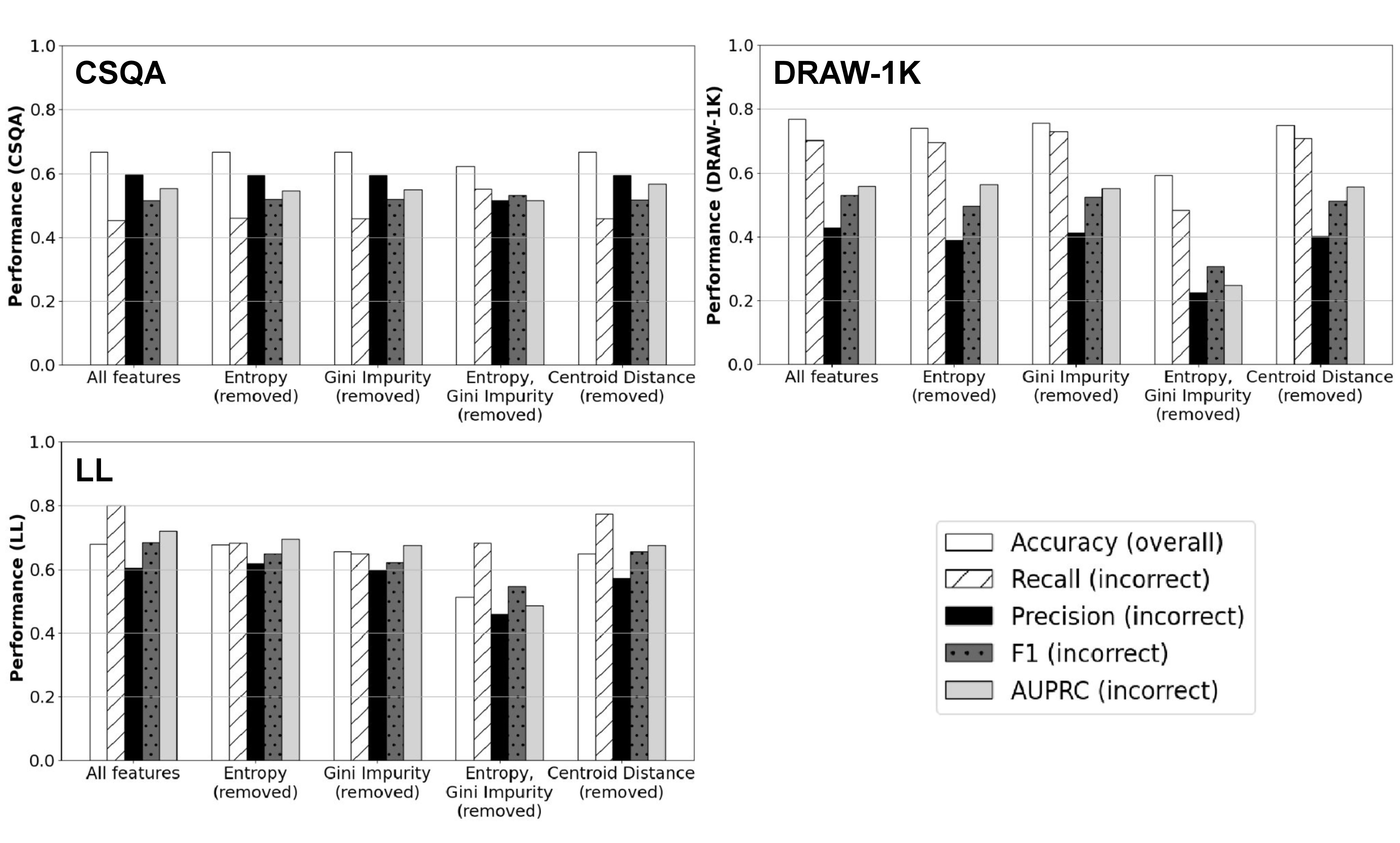}
\caption{Ablation of various diversity measures for the failure prediction model.  Accuracy, precision, recall, F1, and AUPRC are all with respect to predicting LLM failure (MLP-10 for classifier, temperature=0.7).}
\label{fig:ablation}
\end{figure}

\noindent\textbf{Ablation Study.}  We studied the effects of ablating different diversity measures within the MLP-10 model. This model was trained to optimize accuracy in predicting overall performance in the LLM for each of the three data sets.  Results are depicted in Figure~\ref{fig:ablation}. 
 Ablating entropy or Gini impurity separately generally had little effect, which was unsurprising as other experiments in this paper demonstrated that they behave similarly.  However, ablating both results in a significant reduction in performance, most notably with DRAW-1K.

\vspace{5pt}

\noindent\textbf{Alternative Classification Methods.}  We also examined precision and recall results for a variety of classifiers including dense networks for various depths, XGBoost, and AdaBoost. Results for CSQA, DRAW-1K, and LL are shown in Tables~\ref{tab:ec:csqa}, \ref{tab:ec:draw}, and \ref{tab:ec:ll} respectively.


\begin{table}[!ht]
\begin{small}
    \centering
    \begin{tabular}{|l|l|l|l|l|l|}
    \hline
        ~ & ~ & \multicolumn{2}{|c|}{Incorrect} & \multicolumn{2}{|c|}{Correct}   \\ 
        Classifier & Acc. & Prec. & Rec.& Prec. & Rec.  \\ \hline\hline
        MLP 10 & \textbf{0.667} & 0.592 & 0.461 & 0.699 & 0.798  \\ \hline
        MLP 15 & \textbf{0.667} & \textbf{0.593} & 0.461 & 0.699 & \textbf{0.799}  \\ \hline
        XGB & 0.628 & 0.525 & \textbf{0.474} & 0.685 & 0.727  \\ \hline
        AdaBoost & 0.665 & 0.589 & 0.458 & 0.698 & 0.796  \\ \hline
    \end{tabular}
    \caption{CSQA failure prediction results}
    \label{tab:ec:csqa}
\end{small}
\end{table}

\begin{table}[!ht]
    \begin{small}
    \centering
    \begin{tabular}{|l|l|l|l|l|l|}
    \hline
        ~ & ~ & \multicolumn{2}{|c|}{Incorrect} & \multicolumn{2}{|c|}{Correct}   \\ 
        Classifier & Acc. & Prec. & Rec.& Prec. & Rec.  \\ \hline\hline
        MLP 5 & 0.748 & 0.398 & \textbf{0.701} & \textbf{0.920} & 0.759  \\ \hline
        MLP 10 & 0.762 & \textbf{0.414} & 0.690 & 0.919 & 0.778  \\ \hline
        MLP 15 & 0.757 & 0.412 & 0.696 & 0.919 & 0.771  \\ \hline
        XGB & \textbf{0.769} & 0.379 & 0.381 & 0.860 & \textbf{0.857}  \\ \hline
        AdaBoost & 0.741 & 0.380 & 0.636 & 0.904 & 0.765  \\ \hline
    \end{tabular}
    \caption{DRAW-1K failure prediction results}
    \label{tab:ec:draw}
\end{small}
\end{table}

\begin{table}[!ht]
    \begin{small}
        
    \centering
    \begin{tabular}{|l|l|l|l|l|l|}
    \hline
        ~ & ~ & \multicolumn{2}{|c|}{Incorrect} & \multicolumn{2}{|c|}{Correct}   \\ 
        Classifier & Acc. & Prec. & Rec.& Prec. & Rec.  \\ \hline\hline
        MLP 5 & 0.657 & 0.596 & 0.706 & 0.733 & 0.620  \\ \hline
        MLP 10 & 0.657 & 0.613 & \textbf{0.738} & \textbf{0.766} & 0.613  \\ \hline
        MLP 15 & \textbf{0.680} & \textbf{0.637} & 0.656 & 0.730 & \textbf{0.698}  \\ \hline
        XGB & 0.669 & 0.619 & 0.644 & 0.714 & 0.688  \\ \hline
        AdaBoost & 0.678 & 0.614 & 0.734 & 0.756 & 0.635  \\ \hline
    \end{tabular}
\caption{LL failure prediction results}
\label{tab:ec:ll}
\end{small}
    
\end{table}

\section{Conclusion}
In this paper, we presented methods that can be used as a proxy for failure probability in LLM responses to QA tasks without domain-specific or access to the LLM.  The three measures introduced were closely correlated with failure probability across three datasets and we demonstrated several applications including error prediction, prompt engineering, and error correction.  These applications demonstrated how, often without the requirement for supervised data, we could obtain improved results over traditional prompting methods.  Further integration with other techniques such as the recently introduced ``tree of thought'' prompting~\cite{yao_tree_2023} is a direction for future work.  We also believe the application to automatic code generation~\cite{Vaithilingam22}, and the creation of logical sentences~\cite{liu2022langltl} for reasoning tasks (e.g., the rules used in \cite{aditya_pyreason_2023}) is an important application due to the precision required in the output of an LLM when translating natural language to computer code and/or formal logic.  Additionally, diversity measures may also have applications to creative use cases where there is no ``correct'' answer.  In some of these use cases (e.g., for brainstorming in an artistic project) an increase in diversity of response may be desirable.

\section{Acknowledgments}
Some of the authors are supported by internal funding from ASU Fulton Schools of Engineering.

\bibliography{llm_refs,mwp,ds_ref}

\end{document}